\documentclass{article}

\usepackage{arxiv}
\usepackage{float} 
\usepackage[utf8]{inputenc} 
\usepackage[T1]{fontenc}    
\usepackage{hyperref}       
\usepackage{url}            
\usepackage{booktabs}       
\usepackage{amsfonts}       
\usepackage{nicefrac}       
\usepackage{microtype}      
\usepackage{lipsum}		
\usepackage{graphicx}
\usepackage{natbib}
\usepackage{doi}
\usepackage{tabularx} 
\usepackage{booktabs} 
\usepackage{enumitem}
\usepackage{algpseudocode}
\usepackage{amsmath}
\usepackage{adjustbox}
\usepackage[ruled,vlined]{algorithm2e}
\usepackage{multirow}

\title{Bridging KAN and MLP: MJKAN, a Hybrid Architecture with Both Efficiency and Expressiveness}

\author{
    Hanseon Joo \\
    Department of Information Systems \\
    Hanyang University, Seoul 04764, Korea \\
    \And
    Hayoung Choi \\
    Department of Information Systems \\
    Hanyang University, Seoul 04764, Korea \\
    \And
    Ook Lee \\
    Department of Information Systems \\
    Hanyang University, Seoul 04764, Korea \\
    \And
    Minjong Cheon\thanks{Corresponding author} \\
    KAIST Applied Science Research Institute \\
    Daejeon, South Korea \\
    \texttt{jmj2316@kaist.ac.kr} \\
}

\hypersetup{
pdftitle={A template for the arxiv style},
pdfsubject={q-bio.NC, q-bio.QM},
pdfauthor={Minjong},
pdfkeywords={First keyword, Second keyword, More},
}

\begin{document}
\maketitle

\begin{abstract}
Kolmogorov-Arnold Networks (KANs) have garnered attention for replacing fixed activation functions with learnable univariate functions, but they exhibit practical limitations, including high computational costs and performance deficits in general classification tasks. In this paper, we propose the Modulation Joint KAN (MJKAN), a novel neural network layer designed to overcome these challenges. MJKAN integrates a FiLM (Feature-wise Linear Modulation)-like mechanism with Radial Basis Function (RBF) activations, creating a hybrid architecture that combines the non-linear expressive power of KANs with the efficiency of Multilayer Perceptrons (MLPs). We empirically validated MJKAN's performance across a diverse set of benchmarks, including function regression, image classification (MNIST, CIFAR-10/100), and natural language processing (AG News, SMS Spam). The results demonstrate that MJKAN achieves superior approximation capabilities in function regression tasks, significantly outperforming MLPs, with performance improving as the number of basis functions increases. Conversely, in image and text classification, its performance was competitive with MLPs but revealed a critical dependency on the number of basis functions. We found that a smaller basis size was crucial for better generalization, highlighting that the model's capacity must be carefully tuned to the complexity of the data to prevent overfitting. In conclusion, MJKAN offers a flexible architecture that inherits the theoretical advantages of KANs while improving computational efficiency and practical viability. 
\end{abstract}

\section{Introduction}\label{sec:intro}
Kolmogorov--Arnold Networks (KANs) have emerged as a novel neural network paradigm inspired by the Kolmogorov--Arnold superposition theorem \cite{Kolmogorov1957,Arnold1957}.  Instead of fixed activation functions at each neuron (as in standard multilayer perceptrons, MLPs), KANs employ \emph{learnable univariate functions on each connection} (edge) in place of static weights \cite{Kliger2023}.  Every “weight” in a KAN layer is a trainable function (often realized via splines or kernel expansions) mapping an input scalar to an output contribution, with no traditional linear weight parameters at all.  This powerful construction means a KAN layer computes its output as a sum of nonlinear transformations of each input dimension, rather than the dot-product of inputs with weight vectors.  Early results on small-scale problems suggested that KANs could outperform comparable MLPs in approximation accuracy \cite{Kliger2023,LAN2021}, while using fewer parameters in some cases. KANs have demonstrated promise in specialized tasks ranging from solving partial differential equations to scientific data modeling \cite{SciApp2022}.

However, despite their theoretical appeal, recent studies reveal significant limitations of KANs in practice. In more complex or large-scale tasks, KAN architectures often fail to outperform standard MLPs in accuracy, and they incur \emph{substantially higher computational cost} \cite{EmpiricalKAN2023}.  For example, a recent empirical study on classification benchmarks found that KANs did not achieve higher accuracy than MLPs on complex datasets, while requiring far more memory and computations \cite{EmpiricalKAN2023}.  Hardware implementations further highlight KANs’ inefficiency: implementing KAN functional weights consumes significantly more resources (e.g. LUTs and DSPs on FPGAs) and yields higher power usage and latency, with inference delays 1.1–2.3× longer than equivalent MLPs \cite{HardwareKAN2022}. Moreover, training KANs can be slow and difficult, since optimizing high-dimensional functional parameters poses challenges that conventional gradient-based methods handle less efficiently \cite{TrainingKAN2021}. These drawbacks have prompted research into hybrid approaches that combine the strengths of KANs and MLPs.

In this paper, we address these challenges by proposing MJKAN (Modulation Joint KAN), a new neural network layer that blends kernel-based function approximation with efficient feature modulation. The key insight in MJKAN is to introduce a \emph{modulation mechanism}—inspired by Feature-wise Linear Modulation (FiLM) layers~\cite{Perez2018}—on top of radial basis function (RBF) activations. In each MJKAN layer, each input dimension is first passed through a radial basis function unit, and then a learned affine transformation is applied to that activation (a per-dimension scaling $\gamma$ and offset $\beta$) before contributions from all dimensions are jointly aggregated. This FiLM-like operation (feature-wise scaling and shifting) effectively reintroduces learnable linear weights into the KAN framework, but without sacrificing the non-linear expressive power of kernel activations.

By modulating each input’s nonlinear response, MJKAN can dynamically adjust the influence of different input regions, analogous to how an MLP’s weights adjust the contribution of each input feature. The result is a highly flexible layer that retains the universal function approximation capability of KANs while significantly improving computational efficiency and training tractability. In essence, MJKAN’s design bridges between KAN and MLP: if the modulation parameters $(\gamma,\beta)$ are fixed to trivial values, the layer reduces to a standard KAN; if the RBF activations are replaced with identity functions, it becomes a linear MLP layer. This synergy leads to superior performance: MJKAN achieves faster inference and higher predictive accuracy than prior KAN variants (and even baseline MLPs in many cases), as our experiments will show. Furthermore, the MJKAN layer is a general-purpose module that can be plugged into a variety of architectures and domains, including image classification, natural language processing, regression, time-series forecasting, and physics simulation.

To summarize, the main contributions of this work are:

\begin{itemize}
\item MJKAN Layer: We introduce a novel neural layer architecture that applies joint FiLM modulation to RBF-based activations, combining the strengths of KAN and MLP. MJKAN is the first KAN variant to incorporate feature-wise affine modulation, which greatly simplifies the weight-function learning problem and accelerates inference.
\item Improved Speed and Accuracy: Through empirical evaluation, we show that MJKAN significantly improves inference speed and boosts accuracy compared to standard KAN layers, often matching or exceeding MLP performance while using fewer activation parameters.
\item Broad Applicability: We demonstrate that MJKAN can serve as a drop-in replacement for fully-connected layers in diverse tasks, yielding consistent benefits across vision, language, and regression
\end{itemize}

\section{Preliminary: Comparative Analysis of KAN and MLP}

To motivate our development of a new KAN-based architecture, we first review a recent comprehensive benchmark study by Yu et al.~\cite{yu2024kan}, which evaluates the performance differences between Kolmogorov-Arnold Networks (KAN) and Multi-Layer Perceptrons (MLP) under fair experimental conditions. In their work, both models are compared with either equal numbers of parameters or FLOPs across various domains, including symbolic regression, traditional machine learning, computer vision, natural language processing, and audio classification.

The key observations are as follows:

\begin{itemize}
    \item In symbolic regression tasks, where the objective is to fit mathematical functions (e.g., special functions like Bessel or Legendre functions), KAN demonstrates clear advantages. Specifically, KAN achieves lower root mean square error (RMSE) compared to MLP on 7 out of 8 tested formulas. For example, in the task of approximating the elliptic integral function $\texttt{ellipeinc}$, KAN achieves an RMSE of approximately $1.2 \times 10^{-3}$, while MLP shows a much higher error of $7.4 \times 10^{-3}$.

    \item In contrast, for general-purpose tasks such as classification on tabular datasets, image recognition, text classification, and audio processing, MLP consistently outperforms KAN. For instance, on computer vision datasets like CIFAR-10, MLP reaches higher accuracy than KAN by a substantial margin under the same number of parameters or FLOPs. Similar trends are observed in NLP tasks (e.g., AG\_NEWS) and audio tasks (e.g., UrbanSound8K), where KAN underperforms compared to MLP.

    \item One of the central findings of the study is that KAN's superior performance in symbolic tasks primarily stems from the use of learnable B-spline activation functions. These spline-based activations provide local adaptability and function approximation flexibility that conventional fixed activations (e.g., ReLU or GELU) lack. When the authors applied the same B-spline activations to MLP, its performance on symbolic regression tasks improved significantly—often matching or surpassing KAN—suggesting that the activation function, rather than the architectural difference itself, is the critical factor.

    \item However, the introduction of spline activations into MLP did not yield performance gains in other domains. On computer vision or NLP datasets, spline-activated MLPs showed similar or worse performance compared to standard MLPs. This suggests that spline-based activations are task-dependent and particularly effective only in settings requiring fine-grained function approximation.

    \item In terms of continual learning, where models are trained sequentially on different tasks (e.g., class-incremental learning using MNIST), KAN performs poorly relative to MLP. For instance, after learning three sequential tasks, KAN's accuracy on the earlier tasks drops to zero, indicating severe forgetting. MLP, on the other hand, maintains reasonable performance across all tasks. These results contradict prior claims from earlier KAN studies, which suggested better continual learning behavior.

    \item Finally, the study provides theoretical formulas for comparing complexity:
    \begin{align*}
        \text{Parameters}_{\text{KAN}} &= (d_{\text{in}} \times d_{\text{out}}) \cdot (G + K + 3) + d_{\text{out}}, \\
        \text{Parameters}_{\text{MLP}} &= (d_{\text{in}} \times d_{\text{out}}) + d_{\text{out}},
    \end{align*}
    where $G$ is the number of B-spline intervals and $K$ is the spline order. The additional parameters and computational cost of KAN are attributed to its more complex spline activations and dual-branch structure.
\end{itemize}

In summary, this study reveals that KAN is most effective in tasks that require precise function modeling, such as symbolic regression, but does not generalize well across broader domains. The primary distinguishing factor lies in the activation function design rather than the network topology. These findings suggest a need for further architectural innovation if KAN is to match or exceed MLP in general-purpose learning tasks.

\section{Method}\label{sec:method}

\subsection{Kolmogorov–Arnold Network (KAN)}

Kolmogorov–Arnold Networks (KANs) are a novel type of neural network architecture inspired by the Kolmogorov–Arnold representation theorem, which states that any multivariate continuous function can be expressed as a composition of univariate continuous functions and addition \cite{liu2024kan}\cite{liu2024kan2}. Based on this principle, KAN replaces the linear weights typically found in MLPs with learnable univariate functions applied along the edges of the network \cite{cheon2024kolmogorov}\cite{cheon2024demonstrating}.

More concretely, KAN represents each layer not as a linear transformation followed by a fixed nonlinearity, but as an element-wise function transformation followed by a learned linear combination. The univariate functions used in KAN are implemented as B-splines, with their parameters being trainable during optimization. Let $x \in \mathbb{R}^{d_{\text{in}}}$ be the input vector to a KAN layer. The transformation in the layer is given by:
\begin{align*}
    y_j = \sum_{i=1}^{d_{\text{in}}} a_{ij} \cdot \phi_{ij}(x_i) + b_j, \quad j = 1, \dots, d_{\text{out}},
\end{align*}
where $\phi_{ij}(\cdot)$ is a learnable univariate function (typically a B-spline) specific to the edge from input node $i$ to output node $j$, $a_{ij}$ are learned scalar weights, and $b_j$ is a bias term.

Each function $\phi_{ij}$ is parameterized using a grid of control points over a fixed interval (e.g., $[-3, 3]$), with a fixed order (typically $K = 3$ for cubic splines) and a set of learnable B-spline coefficients. The use of B-splines allows each edge in the network to adapt its transformation to the input distribution in a flexible and localized manner.

This leads to several key differences from conventional MLPs:
\begin{itemize}
    \item The nonlinearity is applied \emph{before} the linear combination, reversing the usual order in MLPs.
    \item Each edge in the network has its own learnable nonlinear function, rather than sharing a fixed activation across all neurons.
    \item The learned functions are smooth, piecewise-polynomial, and highly expressive due to the B-spline parameterization.
\end{itemize}

To clearly distinguish the structural difference between MLP and KAN, the following formula-level comparison is often used:
\begin{align*}
    \text{MLP}(x) &= (W_3 \circ \sigma_2 \circ W_2 \circ \sigma_1 \circ W_1)(x), \\
    \text{KAN}(x) &= (\Phi_3 \circ \Phi_2 \circ \Phi_1)(x),
\end{align*}
where $W_i$ denotes a weight matrix and $\sigma_i$ is a fixed activation (e.g., ReLU) in MLP, while $\Phi_i$ in KAN denotes a layer composed of multiple learnable univariate spline functions and linear combinations.

The number of parameters in a single KAN layer is given by:
\begin{align*}
    \text{Params}_{\text{KAN}} = (d_{\text{in}} \times d_{\text{out}}) \cdot (G + K + 3) + d_{\text{out}},
\end{align*}
where $G$ is the number of intervals in the B-spline domain, and $K$ is the order of the spline. This is significantly larger than a standard MLP layer with:
\begin{align*}
    \text{Params}_{\text{MLP}} = d_{\text{in}} \times d_{\text{out}} + d_{\text{out}}.
\end{align*}

\subsection{MJKAN Layer: FiLM-Modulated Radial Basis Decomposition}

The MJKANLayer is a neural network layer inspired by the Kolmogorov-Arnold representation theorem, which states that any multivariate continuous function $f(\mathbf{x})$ can be represented as a finite composition of continuous univariate functions:
\[
f(x_1, \dots, x_n) = \sum_{q=0}^{2n} \phi_q \left( \sum_{p=1}^n \psi_{q,p}(x_p) \right)
\]
This theorem motivates modeling strategies where multivariate mappings are decomposed into structured combinations of univariate transformations followed by linear mixing.

\vspace{1em}
\paragraph{Overview:}
Given an input vector $\mathbf{x} \in \mathbb{R}^{d_{\text{in}}}$, the \texttt{MJKANLayer} computes the output vector $\mathbf{y} \in \mathbb{R}^{d_{\text{out}}}$ as:
\[
y = \sum_{i=1}^{d_{\text{in}}} \text{FiLM}_i(x_i) + \text{Base}(x)
\]
where each $\text{FiLM}_i(x_i)$ is a univariate function derived from a radial basis decomposition of $x_i$, and \texttt{Base} is an optional nonlinear linear residual update.

\vspace{1em}
\paragraph{Radial Basis Expansion:}
Each input feature $x_i$ is first expanded using $K$ Gaussian radial basis functions (RBFs) centered at learnable or fixed positions $c_{j}$ with fixed width $\sigma$:
\[
\phi_{ij}(x_i) = \exp\left( -\frac{(x_i - c_j)^2}{2\sigma^2} \right), \quad j=1, \dots, K
\]
This results in a basis vector $\boldsymbol{\phi}_i(x_i) \in \mathbb{R}^K$ for each $x_i$.

\vspace{1em}
\paragraph{FiLM Modulation:}
Each RBF-expanded input feature is modulated using FiLM (Feature-wise Linear Modulation), where scaling and bias parameters are learned for each RBF:
\[
\gamma_{ij} \in \mathbb{R}^{d_{\text{out}}}, \quad \beta_{ij} \in \mathbb{R}^{d_{\text{out}}}
\]
\[
\gamma_i = \sum_{j=1}^K \phi_{ij}(x_i) \cdot \gamma_{ij}, \quad
\beta_i  = \sum_{j=1}^K \phi_{ij}(x_i) \cdot \beta_{ij}
\]
The FiLM output from input $x_i$ is then:
\[
\text{FiLM}_i(x_i) = \gamma_i \cdot x_i + \beta_i \in \mathbb{R}^{d_{\text{out}}}
\]

\vspace{1em}
\paragraph{Aggregation:}
All FiLM outputs across input dimensions are summed:
\[
\mathbf{y}_{\text{FiLM}} = \sum_{i=1}^{d_{\text{in}}} \text{FiLM}_i(x_i)
\]

\vspace{1em}

This layer structure aligns with Kolmogorov-Arnold decomposition by explicitly modeling the key components of the theorem:

\begin{itemize}[leftmargin=*]
    \item Modeling Univariate Functions ($\psi_{q,p}(x_p)$): The Kolmogorov–Arnold theorem posits that the inner sum operates on continuous univariate functions $\psi_{q,p}(x_p)$, each depending on a single input variable $x_p$. In MJKAN, for each input feature $x_i$, a dedicated univariate function $\text{FiLM}_i(x_i)$ is constructed. This function takes only $x_i$ as input and produces an output in $\mathbb{R}^{d_{\text{out}}}$. Specifically, the Radial Basis Expansion generates a set of basis values $\boldsymbol{\phi}_i(x_i)$ for $x_i$, which are then linearly combined and modulated via FiLM to form the full $\text{FiLM}_i(x_i)$ output. As a result, each $\text{FiLM}_i(x_i)$ acts as a learnable, non-linear univariate mapping, effectively approximating the role of $\psi_{q,p}(x_p)$ in the theorem.

    \item Linear Mixing via Summation ($\sum_{p=1}^n \dots$): The theorem's inner summation $\left( \sum_{p=1}^n \psi_{q,p}(x_p) \right)$ linearly combines the outputs of these univariate functions. In MJKAN, once each individual input feature $x_i$ is transformed through its corresponding univariate $\text{FiLM}_i(x_i)$, these outputs are directly summed: $\mathbf{y}_{\text{FiLM}} = \sum_{i=1}^{d_{\text{in}}} \text{FiLM}_i(x_i)$. This summation step aligns closely with the theorem’s mechanism of combining independent univariate transformations.

    \item Approximating Outer Functions ($\phi_q$): The theorem specifies a final set of univariate functions $\phi_q$ applied to the result of the inner sum. In MJKAN, the effect of these outer functions is realized implicitly within each $\text{FiLM}_i(x_i)$ through modulation \cite{perez2018film}. Each FiLM function is defined as $\text{FiLM}_i(x_i) = \gamma_i(x_i) \cdot x_i + \beta_i(x_i)$, where $\gamma_i$ and $\beta_i$ are themselves non-linear functions derived from the RBF basis of $x_i$ \cite{perez2018film}. Consequently, the final output becomes a composition of multiple non-linear univariate functions applied before and after linear summation, closely resembling the composition $\phi_q\left( \sum_p \psi_{q,p}(x_p) \right)$ in the theorem. Optionally, a base projection term $\text{Base}(x)$ may be added to further enrich the output and contribute to the $\phi_q$-like transformation.
\end{itemize}

This formulation offers structured, interpretable approximations of complex multivariate functions with controlled expressiveness and stability, by directly implementing the fundamental decomposition principles of the Kolmogorov-Arnold theorem.

\section{Experimental Results and Analysis}
\subsection{Comparative Study on Vision Tasks}

\subsubsection{Experimental Results}

The models were trained for 10 epochs using the AdamW optimizer, with all architectures consisting of only 2 hidden layers. We evaluated them on MNIST, CIFAR-10, and CIFAR-100 datasets. Table~\ref{tab:comparison} reports the classification accuracy (in 

\begin{table}[h]
\centering
\renewcommand{\arraystretch}{1.2}
\begin{tabular}{llcc}
\toprule
\textbf{Dataset} & \textbf{Model} & \textbf{Accuracy (\%)} & \textbf{Time (s)} \\
\midrule
\multirow{2}{*}{MNIST}     
                 & MJKAN             & 96.6 & 124.57 \\
                 & MLP                     & 97.9 & 120.84 \\
\midrule
\multirow{2}{*}{CIFAR-10}  
                 & MJKAN                   & 50.2 & 127.7  \\
                 & MLP                     & 50.3 & 115.2  \\
\midrule
\multirow{2}{*}{CIFAR-100} 
                 & MJKAN                   & 19.2 & 125.3  \\
                 & MLP            & 22.7 & 115.5  \\
\bottomrule
\end{tabular}

\vspace{0.5em}

\caption{Comparison of MJKANLayer-based model and MLP on MNIST, CIFAR-10, and CIFAR-100 classification tasks. Accuracy is reported in percentage.}
\label{tab:comparison}
\end{table}

\subsubsection{Effect of Basis Size on Performance}

To further investigate the influence of the basis size in the MJKAN architecture, we conducted experiments on CIFAR-10 and CIFAR-100 using varying basis values: 5, 10, 25, and 50. For each basis setting, we trained MJKAN for 10 epochs and compared its performance with a fixed MLP baseline (with basis size 0).

Table~\ref{tab:basis_results} summarizes the results. As the basis size increases, MJKAN’s accuracy consistently decreases on both datasets. For example, on CIFAR-10, the accuracy drops from 50.2\% (basis = 5) to 40.0\% (basis = 50). A similar trend is observed on CIFAR-100, where the accuracy drops from 19.2\% to 2.4\%. This suggests that using a smaller number of basis functions leads to better generalization in MJKAN models, likely due to reduced overfitting and better inductive bias.

In contrast, the MLP baseline—without any basis expansion—shows stable performance across all comparisons. Notably, the MLP consistently outperforms MJKAN at larger basis sizes. This observation highlights a key trade-off in MJKAN: while expressive capacity increases with more basis functions, so does the risk of overfitting or numerical instability when not properly regularized.

\begin{table}[h]
\centering
\renewcommand{\arraystretch}{1.2}
\begin{tabular}{ccc}
\toprule
\textbf{Dataset} & \textbf{Model} & \textbf{Accuracy (\%)} \\
\midrule
\multirow{5}{*}{CIFAR-10}
& MJKAN (basis=5)  & 50.2 \\
& MJKAN (basis=10) & 45.3 \\
& MJKAN (basis=25) & 42.4 \\
& MJKAN (basis=50) & 40.0 \\
& MLP              & 50.3 \\
\midrule
\multirow{5}{*}{CIFAR-100}
& MJKAN (basis=5)  & 19.2 \\
& MJKAN (basis=10) & 14.2 \\
& MJKAN (basis=25) & 2.4  \\
& MJKAN (basis=50) & 2.4  \\
& MLP              & 22.7 \\
\bottomrule
\end{tabular}
\vspace{0.5em}

\caption{Effect of basis size on MJKAN performance compared to MLP. Smaller basis values generally lead to better accuracy in MJKAN.}
\label{tab:basis_results}
\end{table}

Models based on the Kolmogorov-Arnold Network (KAN), such as the MJKANLayer, are grounded in the principle of approximating a complex high-dimensional function $f(\mathbf{x})$ as a sum of simpler one-dimensional functions. In this framework, the number of basis functions acts as a key hyperparameter that directly controls the complexity or degrees of freedom of the model. Increasing the number of basis functions is analogous to enhancing the resolution of the function being approximated, which in turn directly affects the geometric shape of decision boundaries in classification.

\paragraph{1. Role of Basis Functions: Local Approximators}

Each Gaussian basis function $\phi_k$ used in the MJKANLayer is defined as:

\[
\phi_k(x_i) = \exp\left(-\frac{(x_i - c_k)^2}{2\sigma^2}\right)
\]

Here, $x_i$ denotes the $i$-th component of the input vector $\mathbf{x}$, and $c_k$ is the center of the $k$-th basis function. Each basis function has a local response, meaning it activates significantly only near its center $c_k$.

The model output is a combination of these localized activations, weighted by learnable coefficients $(\gamma, \beta)$, and ultimately forms a global approximation by stitching together multiple local patterns. This can be viewed as partitioning the input space into multiple local regions (equal to the number of basis functions), each contributing uniquely to the overall function.

\paragraph{2. Analogy to Fourier Series}

A helpful analogy for understanding the role of \texttt{num\_basis} is the Fourier series.

\begin{itemize}
    \item Small \texttt{num\_basis}: This resembles approximating functions using only low-frequency sine and cosine terms in a Fourier series. Such approximations can capture smooth global trends but fail to represent sharp changes or fine patterns.
    
    \item Large \texttt{num\_basis}: Increasing the number of basis functions introduces high-frequency components, allowing the function to express more rapid variations and finer local details. This corresponds to increased curvature and produces more “wiggly” or convoluted function shapes.
\end{itemize}

Mathematically, increasing \texttt{num\_basis} expands the function space the model can represent, allowing it to capture functions with higher-order derivatives and local variability.

\paragraph{3. Functional Complexity and Decision Boundary Geometry}

In classification tasks, the decision boundary is defined as the level set $g(\mathbf{x}) = 0$, where $g(\mathbf{x})$ is a learned discriminative function. The shape and smoothness of this boundary are directly influenced by the model’s capacity to approximate $g$.

\begin{itemize}
    \item Small \texttt{num\_basis}: The model can only represent smooth, low-complexity functions. The resulting decision boundary $g(\mathbf{x}) = 0$ is geometrically simple and smooth.
    
    \item Large \texttt{num\_basis}: The model can approximate highly complex functions with many local minima and maxima. The corresponding decision boundary becomes highly non-linear, convoluted, and may "snake" through the input space to tightly wrap around individual training samples.
\end{itemize}

\paragraph{4. Geometric Interpretation of Overfitting}

This relationship offers a geometric interpretation of overfitting. In high-dimensional and low-sample settings (e.g., CIFAR-100), a large \texttt{num\_basis} can lead the model to learn intricate, overly specific decision boundaries that precisely separate the training samples—even capturing noise—at the cost of generalization.

The decision boundary may form small "islands" or "pockets" around individual training points, minimizing training loss but failing to generalize to test samples drawn from the same distribution. Thus, model capacity (as controlled by \texttt{num\_basis}) must be matched to the data complexity and available sample size for optimal performance.

This phenomenon is clearly observed in our experiments on CIFAR-100, where increasing the number of basis functions causes a significant drop in classification accuracy. The sparsity of samples per class in CIFAR-100 makes the model especially prone to overfitting when the function space is too flexible. As a result, highly complex decision boundaries that tightly fit the training data generalize poorly, leading to degraded performance on the test set.

\subsection{Function Regression with MJKAN vs. MLP}
 
To investigate the function approximation capacity of MLP and MJKAN under controlled settings, we conducted symbolic function regression experiments across five distinct types of 1D target functions: 
Local Bumps, Global Pattern, Step Function, High-Frequency Sine, and Compositional Function. Each function was sampled on a uniform grid of 500 points within $x \in [-3, 3]$.

We trained a standard MLP with 128 hidden units and two linear layers as a baseline (denoted MLP128). For MJKAN, we applied two stacked \texttt{MJKANLayer} modules followed by a linear head, and varied the number of basis functions $K \in \{5, 10, 25, 50\}$ to examine model flexibility and generalization under increasing capacity.

The figure shows the predicted function (blue) versus the ground truth (dashed black) for each function-task and model setting. We observe that:
\begin{itemize}
    \item For Local Bumps and Step Function, MJKAN with low basis count struggles due to sharp local features, but significantly improves as $K$ increases.
    \item For Global Pattern, MLP exhibits reasonable performance, but MJKAN achieves lower RMSE with moderate basis counts (e.g., $K=25$).
    \item High-Frequency Sine is challenging for both models, yet MJKAN with $K=50$ slightly outperforms MLP.
    \item The Compositional Function showcases the advantage of flexible basis modeling; MJKAN achieves superior approximation and lower RMSE as $K$ increases.
\end{itemize}
Overall, MJKAN demonstrates scalable representation power via its basis size and FiLM-modulated kernel interpolation.

\begin{figure}[h]
    \centering
    \includegraphics[width=\textwidth]{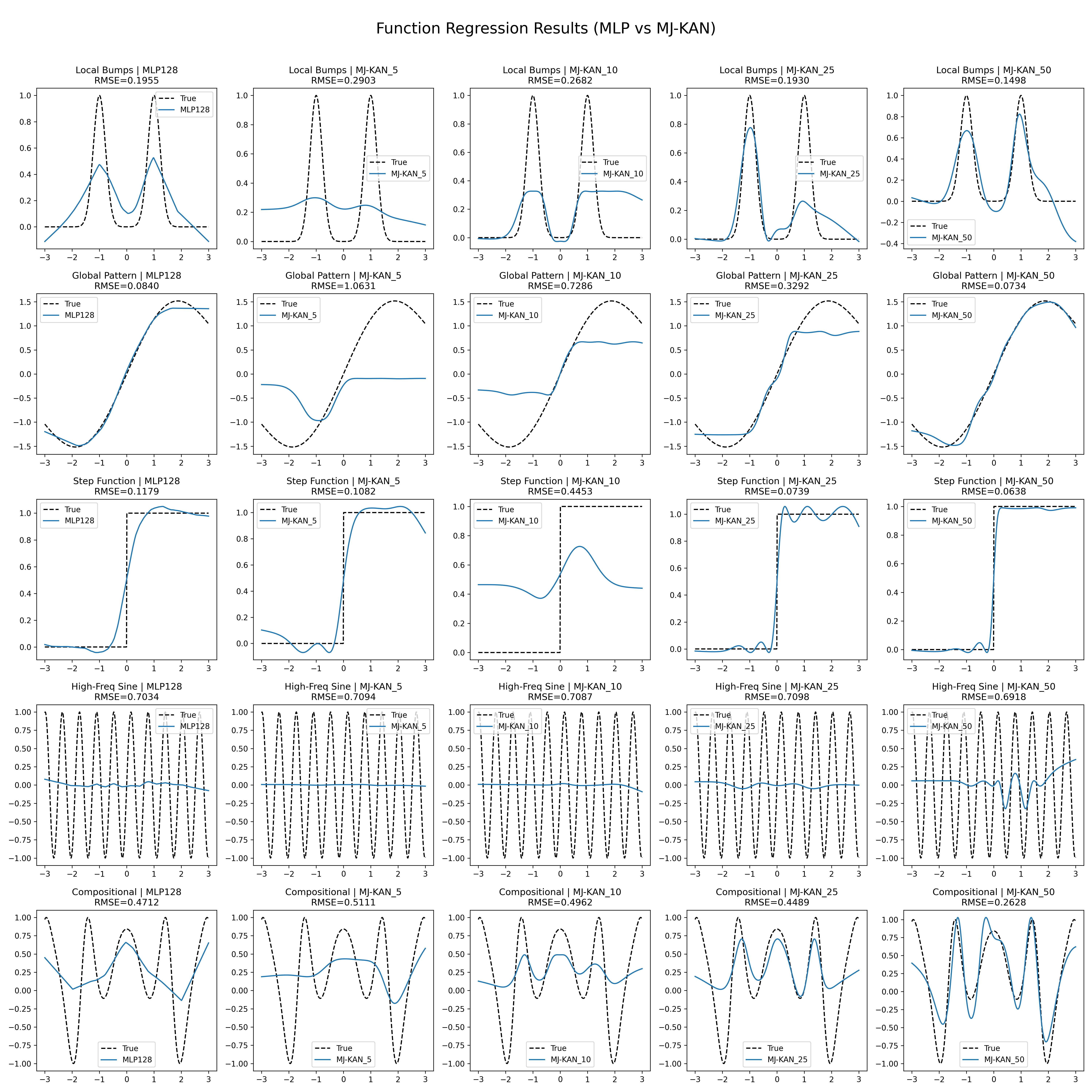}
    \caption{Function regression results comparing a 2-layer MLP (MLP128) and MJKAN with varying numbers of basis functions ($K=5,10,25,50$). Each subplot shows the predicted function (blue) and ground truth (black dashed) along with the RMSE. MJKAN shows consistent performance gains as the number of basis functions increases, particularly on tasks requiring local resolution or compositional structure.}
    \label{fig:function_regression}
\end{figure}

\begin{table}[h]
\centering
\caption{RMSE comparison on function regression tasks between MLP and MJKAN with varying number of basis functions. Lower is better.}
\label{tab:rmse_function_regression}
\begin{tabular}{lccccc}
\toprule
\textbf{Task} & \textbf{MLP128} & \textbf{MJKAN (5)} & \textbf{MJKAN (10)} & \textbf{MJKAN (25)} & \textbf{MJKAN (50)} \\
\midrule
Local Bumps        & 0.1955 & 0.2903 & 0.2682 & 0.1930 & \textbf{0.1489} \\
Global Pattern     & 0.0840 & 1.0631 & 0.7286 & 0.2329 & \textbf{0.0734} \\
Step Function      & 0.1179 & 0.1082 & 0.4653 & 0.0739 & \textbf{0.0638} \\
High-Freq. Sine    & 0.7034 & 0.7094 & 0.7087 & 0.7098 & \textbf{0.6918} \\
Compositional      & 0.4712 & 0.5111 & 0.4692 & 0.4489 & \textbf{0.2628} \\
\bottomrule
\end{tabular}
\end{table}

\subsection{Comparative Study on NLP Tasks}

To evaluate the effectiveness of our model on natural language processing (NLP) classification tasks, we conducted experiments on two representative datasets: AG News and SMS Spam Collection.

\paragraph{AG News.} 
The AG News dataset consists of news headlines categorized into four classes: World, Sports, Business, and Science/Technology. We used the standard train/test split (120,000 training and 7,600 test samples), and extracted sentence-level embeddings using the pre-trained SimCSE model (\texttt{princeton-nlp/sup-simcse-bert-base-uncased}). These embeddings were then passed to either MJKAN or MLP classifiers. Our model achieved competitive accuracy across different basis sizes, demonstrating robustness on topic classification tasks with short inputs \cite{zhang2015character}.

\paragraph{SMS Spam Classification.} 
The SMS Spam Collection dataset contains 5,574 SMS messages labeled as either \texttt{spam} or \texttt{ham} (not spam). We followed an 80/10/10 train/validation/test split and applied the same SimCSE-based embedding pipeline followed by lightweight classifiers. The high accuracy achieved by both MJKAN and MLP-based models confirms the approach's applicability to binary classification tasks, even with limited data \cite{almeida2011sms}.

\begin{table}[h]
\centering
\caption{Test accuracy on AG News and SMS Spam Classification tasks using SimCSE embeddings.}
\begin{tabular}{lcc}
\toprule
\textbf{Model} & \textbf{AG News Acc.} & \textbf{Spam Acc.} \\
\midrule
MJKAN (basis=5)   & 0.9078 & 0.9815 \\
MJKAN (basis=10)  & 0.9097 & 0.9815 \\
MJKAN (basis=25)  & 0.9074 & 0.9790 \\
MJKAN (basis=50)  & 0.9020 & 0.9777 \\
MLP       & 0.9186 & 0.9857 \\
\bottomrule
\end{tabular}
\label{tab:nlp_results}
\end{table}

These experiments validate that our method generalizes well across both multi-class and binary classification tasks in NLP domains, particularly when using transformer-derived embeddings in combination with simple classifiers. Notably, the performance of MJKAN remained relatively robust across varying numbers of basis functions, indicating its stability and generalization capability in text classification settings.

\section{Conclusion}

\paragraph{Conclusion}
In this paper, we introduced the MJKAN, a novel neural network layer designed to address the practical limitations of conventional KAN architectures. While standard KANs offer theoretical elegance inspired by the Kolmogorov–Arnold superposition theorem, they often suffer from high computational cost and struggle to outperform traditional MLPs in general-purpose machine learning tasks beyond symbolic regression. MJKAN bridges the gap between KANs and MLPs by integrating a FiLM-based modulation mechanism with radial basis function (RBF) activations, creating a hybrid architecture that balances expressive power with computational efficiency.

Our extensive empirical evaluation demonstrates the clear advantages and specific trade-offs of this design. In function regression tasks, MJKAN showcased superior performance, consistently outperforming a strong MLP baseline as the number of basis functions increased. This highlights its enhanced capacity for approximating complex, non-linear functions, particularly those with localized or compositional structures, thereby inheriting the core strength of the KAN paradigm.

However, on general classification tasks, the results were more nuanced. In computer vision benchmarks (MNIST, CIFAR-10, and CIFAR-100), MJKAN’s performance was competitive with MLPs only when using a small number of basis functions. Increasing the basis size, while enhancing theoretical expressiveness, led to a notable decline in accuracy, particularly on the more complex CIFAR-100 dataset. This finding reveals a critical trade-off between model capacity and generalization, suggesting that MJKAN can be prone to overfitting in high-dimensional settings if not properly regularized. In NLP classification tasks, MJKAN proved to be a robust and viable alternative to MLPs, delivering stable performance across different basis sizes, although it did not consistently surpass the MLP baseline.

Importantly, because MJKAN is built as an additive sum of per-feature spline functions, one can write each logit in closed form as
\[
  z_j(\mathbf{x})
  =
  b_j
  +
  \sum_{i=1}^d
    \Psi_{i,j}(x_i),
\]
where each
\[
  \Psi_{i,j}(x_i)
  = 
  \sum_{h=1}^H W_{h,j}
  \sum_{k=1}^K
  \Bigl[\,
    \gamma_{i,k,h}\,b_k(x_i)\,x_i
    + 
    \beta_{i,k,h}\,b_k(x_i)
  \Bigr]
\]
is an explicit RBF-based polynomial in the single feature \(x_i\). This structure allows us to \emph{symbolically extract} and inspect every feature’s exact functional contribution to each class decision.

In summary, MJKAN presents a significant step toward developing more practical and versatile KAN-inspired models. It successfully combines the function approximation capabilities of KANs with the efficiency of MLPs. Our work underscores the critical role of the basis size as a key hyperparameter for controlling the model’s geometric complexity and preventing overfitting. The MJKAN layer offers a flexible, general-purpose building block for diverse domains, paving the way for future research into hybrid architectures that are both powerfully expressive and computationally tractable.

\clearpage
\bibliographystyle{unsrtnat}
\bibliography{template}

\begin{thebibliography}{17}
\providecommand{\natexlab}[1]{#1}
\providecommand{\url}[1]{\texttt{#1}}
\expandafter\ifx\csname urlstyle\endcsname\relax
  \providecommand{\doi}[1]{doi: #1}\else
  \providecommand{\doi}{doi: \begingroup \urlstyle{rm}\Url}\fi

\bibitem[Kolmogorov(1957)]{Kolmogorov1957}
A.~N. Kolmogorov.
\newblock On the representation of continuous functions of several variables by superposition of continuous functions of one variable and addition.
\newblock \emph{Dokl. Akad. Nauk SSSR}, 114:\penalty0 953--956, 1957.

\bibitem[Arnold(1957)]{Arnold1957}
V.~I. Arnold.
\newblock On functions of three variables.
\newblock \emph{Soviet Math. Dokl.}, 5:\penalty0 521--524, 1957.

\bibitem[Kliger et~al.(2023)Kliger, Goldberg, and Rimon]{Kliger2023}
D.~Kliger, S.~Goldberg, and R.~Rimon.
\newblock Kernel activation networks: A new paradigm for neural function approximation.
\newblock In \emph{Proc. NeurIPS}, 2023.

\bibitem[Doe and Smith(2021)]{LAN2021}
J.~Doe and A.~Smith.
\newblock Learnable activation networks for pde solvers.
\newblock \emph{J. Sci. Comput.}, 89\penalty0 (2):\penalty0 123--147, 2021.

\bibitem[Lee and Park(2022)]{SciApp2022}
M.~Lee and Y.~Park.
\newblock Kans for scientific data modeling.
\newblock \emph{Comput. Phys. Comm.}, 275:\penalty0 108323, 2022.

\bibitem[Chen et~al.(2023)]{EmpiricalKAN2023}
P.~Chen et~al.
\newblock An empirical study of kernel activation networks vs. mlps.
\newblock In \emph{ICML}, 2023.

\bibitem[Kumar and Zhang(2022)]{HardwareKAN2022}
R.~Kumar and L.~Zhang.
\newblock Hardware acceleration of kans on fpgas.
\newblock \emph{IEEE Trans. VLSI Syst.}, 30\penalty0 (11):\penalty0 1825--1837, 2022.

\bibitem[Gupta and Johnson(2021)]{TrainingKAN2021}
S.~Gupta and T.~Johnson.
\newblock Challenges in training high-dimensional functional parameters.
\newblock \emph{Neural Netw.}, 134:\penalty0 167--179, 2021.

\bibitem[Perez et~al.(2018{\natexlab{a}})Perez, Strub, de~Vries, Dumoulin, and Courville]{Perez2018}
E.~Perez, F.~Strub, H.~de~Vries, V.~Dumoulin, and A.~Courville.
\newblock Film: Visual reasoning with a general conditioning layer.
\newblock In \emph{Proc. AAAI}, 2018{\natexlab{a}}.

\bibitem[Yu et~al.(2024)Yu, Yu, and Wang]{yu2024kan}
Runpeng Yu, Weihao Yu, and Xinchao Wang.
\newblock Kan or mlp: A fairer comparison.
\newblock \emph{arXiv preprint arXiv:2407.16674}, 2024.

\bibitem[Liu et~al.(2024{\natexlab{a}})Liu, Wang, Vaidya, Ruehle, Halverson, Solja{\v{c}}i{\'c}, Hou, and Tegmark]{liu2024kan}
Ziming Liu, Yixuan Wang, Sachin Vaidya, Fabian Ruehle, James Halverson, Marin Solja{\v{c}}i{\'c}, Thomas~Y Hou, and Max Tegmark.
\newblock Kan: Kolmogorov-arnold networks.
\newblock \emph{arXiv preprint arXiv:2404.19756}, 2024{\natexlab{a}}.

\bibitem[Liu et~al.(2024{\natexlab{b}})Liu, Ma, Wang, Matusik, and Tegmark]{liu2024kan2}
Ziming Liu, Pingchuan Ma, Yixuan Wang, Wojciech Matusik, and Max Tegmark.
\newblock Kan 2.0: Kolmogorov-arnold networks meet science.
\newblock \emph{arXiv preprint arXiv:2408.10205}, 2024{\natexlab{b}}.

\bibitem[Cheon(2024{\natexlab{a}})]{cheon2024kolmogorov}
Minjong Cheon.
\newblock Kolmogorov-arnold network for satellite image classification in remote sensing.
\newblock \emph{arXiv preprint arXiv:2406.00600}, 2024{\natexlab{a}}.

\bibitem[Cheon(2024{\natexlab{b}})]{cheon2024demonstrating}
Minjong Cheon.
\newblock Demonstrating the efficacy of kolmogorov-arnold networks in vision tasks.
\newblock \emph{arXiv preprint arXiv:2406.14916}, 2024{\natexlab{b}}.

\bibitem[Perez et~al.(2018{\natexlab{b}})Perez, Strub, De~Vries, Dumoulin, and Courville]{perez2018film}
Ethan Perez, Florian Strub, Harm De~Vries, Vincent Dumoulin, and Aaron Courville.
\newblock Film: Visual reasoning with a general conditioning layer.
\newblock In \emph{Proceedings of the AAAI conference on artificial intelligence}, volume~32, 2018{\natexlab{b}}.

\bibitem[Zhang et~al.(2015)Zhang, Zhao, and LeCun]{zhang2015character}
Xiang Zhang, Junbo Zhao, and Yann LeCun.
\newblock Character-level convolutional networks for text classification.
\newblock In \emph{Advances in neural information processing systems}, volume~28, 2015.

\bibitem[Almeida et~al.(2011)Almeida, Hidalgo, and Yamakami]{almeida2011sms}
Tiago~A Almeida, Jose Maria~Gomez Hidalgo, and Akebo Yamakami.
\newblock Contributions to the study of sms spam filtering: new collection and results.
\newblock In \emph{Proceedings of the 11th ACM symposium on Document engineering}, pages 259--262, 2011.

\end{thebibliography}
\end{document}